\documentclass[letterpaper, 10 pt, conference]{ieeeconf}  % Comment this line out if you need a4paper

\IEEEoverridecommandlockouts                              % This command is only needed if 
\usepackage{graphicx} % for pdf, bitmapped graphics files
\usepackage{amsmath} % assumes amsmath package installed
\usepackage{amssymb}  % assumes amsmath package installed
\usepackage{hyperref}

\title{\LARGE \bf
Enhancing SUMO simulator for simulation based testing and validation of autonomous vehicles
}

\author{Arpan Kusari$^{1}$, Pei Li$^{1}$, Hanzhi Yang$^{2}$, Nikhil Punshi$^{3}$, Mich Rasulis$^{1}$, Scott Bogard$^{1}$ and David J. LeBlanc$^{1}$% <-this % stops a space
\thanks{$^{1}$ Arpan Kusari, Pei Li, Mich Rasulis, Scott Bogard and David J. LeBlanc are with the University of Michigan Transportation Research Institute, University of Michigan, Ann Arbor, MI 48109, USA
        {\tt\small {kusari, peiliesg, mich, sbogard, leblanc}@umich.edu}}%
\thanks{$^{2}$ Hanzhi Yang is with the Department of Mechanical Engineering, University of Michigan, Ann Arbor, MI 48109, USA
        {\tt\small yanghz@umich.edu}}%
\thanks{$^{3}$ Nikhil Punshi is with the Department of Aerospace Engineering, University of Michigan, Ann Arbor, MI 48109, USA
        {\tt\small npunshi@umich.edu}}%
}

\begin{document}

\maketitle
\thispagestyle{empty}
\pagestyle{empty}

%%%%%%%%%%%%%%%%%%%%%%%%%%%%%%%%%%%%%%%%%%%%%%%%%%%%%%%%%%%%%%%%%%%%%%%%%%%%%%%%
\begin{abstract}

Current autonomous vehicle (AV) simulators are built to provide large-scale testing required to prove capabilities under varied conditions in controlled, repeatable fashion. However, they have certain failings including the need for user expertise and complex inconvenient tutorials for customized scenario creation. Simulation of Urban Mobility (SUMO) simulator, which has been presented as an open-source AV simulator, is used extensively but suffer from similar issues which make it difficult for entry-level practitioners to utilize the simulator without significant time investment. In that regard, we provide two enhancements to SUMO simulator geared towards massively improving user experience and providing real-life like variability for surrounding traffic. Firstly, we calibrate a car-following model, Intelligent Driver Model (IDM), for highway and urban naturalistic driving data and sample automatically from the parameter distributions to create the background vehicles. Secondly, we combine SUMO with OpenAI gym, creating a Python package which can run simulations based on real world highway and urban layouts with generic output observations and input actions that can be processed via any AV pipeline. Our aim through these enhancements is to provide an easy-to-use platform which can be readily used for AV testing and validation. 

\end{abstract}

%%%%%%%%%%%%%%%%%%%%%%%%%%%%%%%%%%%%%%%%%%%%%%%%%%%%%%%%%%%%%%%%%%%%%%%%%%%%%%%%
\section{INTRODUCTION}

% why is large scale simulation based testing essential for autonomous vehicles?
With the rise in research into autonomous vehicles (AVs), there has been an increased focus on the effective testing and validation of AVs. Traditional testing and validation for vehicles is performed via repeated closed track testing and open road testing. However, researchers have shown that for AVs, the vehicles would need to be driven hundreds of millions to hundreds of billions of miles to demonstrate reliability which would make AV market penetration improbable \cite{kalra2016driving}. Large scale simulation based testing provides a lucrative alternative for fast and efficient prototyping without the need for a physical vehicle and actual testing in a real world \cite{thorn2018framework}. There are a lot of advantages offered by simulation based testing mostly in terms of controllability, repeatability and scalability which can make it essential for testing highly complex systems such as AVs. As a result, there are various commercial and open-source driving simulation platforms that have been developed for AV testing such as CARLA, Metamoto, Airsim etc. While these simulation software contain sophisticated, intelligent behavior models for vehicular traffic, bicyclists and pedestrians, there are two inherent problems for those who want to utilize these for large scale AV testing and validation:
\begin{itemize}
    \item most of these simulation environments are set up for small scale, repeatable scenarios often in urban layouts which constrain the number of scenarios that can be reliably tested and
    \item they often have a complicated interface which requires a lot of man-hours to learn and run with the AV pipeline in a modular fashion.
\end{itemize}

Simulation of Urban Mobility (SUMO) has been proposed as an alternative for testing autonomous vehicles in their simulation environment \cite{lu2020impact}. SUMO is currently the most popular 2D microscopic traffic simulator which can simulate multimodal traffic, import large scale real-world maps from mapping platforms such as OpenStreetMap (OSM) and can be interacted with using an socket based communication interface, TraCI \cite{krajzewicz2012recent}. The main advantages in utilizing SUMO for large scale AV testing would be access to a large variety of Operational Design Domains (ODDs) that can help AV developers to find edge cases in the pipeline and native support for different car-following models which can provide real-world like behavior of the background vehicles (BVs).  

% What is the motivation for our research?
However, there are still some gaps in using SUMO simulator as a reliable testing environment for AVs:
\begin{itemize}
    \item the BVs behave in a predetermined fashion which is inconsistent with the real-world variability and
    \item there is a steep learning curve in running SUMO and setting it up to interact with the AV pipeline.
\end{itemize}
We propose two enhancements to the SUMO simulator intended to address these two issues. The first enhancement is to learn the parameter distributions of well-known car-following models from naturalistic driving databases and randomly sampling from them to simulate BV behavior. The second enhancement is in providing scenario abstraction by combining the strengths of SUMO simulator with that of OpenAI gym, a very popular and highly accessible toolkit for reinforcement learning research. Our focus on offering these enhancements is on convenience and accessibility while offering real-world like variability. 

There has been some work done in combining OpenAI Gym with driving simulators, one specific popular instance being the highway-env \cite{highway-env} environment. Highway-env is a Python based simulation environment which is built on top of OpenAI gym and contains different environments such as multilane highway, merge, roundabout etc. However, building a custom environment is difficult to do from scratch and the simulation does not scale well with the increase in vehicles and road network layout. Also the background vehicles are sampled from the same parameter values and therefore, offer almost no variability. 

The paper is set up as follows: Section \ref{sec:bayesian} provides the background for estimating parameters of car-following models using Bayesian parameter estimation from naturalistic data and showcases results from in-depth analysis of large amounts of data collected in highway and urban scenarios; Section \ref{sec:scenario}  explores in detail the combination of the SUMO simulator with the OpenAI gym to create a scenario abstraction pipeline and finally, Section \ref{sec:discussion} provides in-depth discussion and some paths towards future work. 

\section{BAYESIAN PARAMETER ESTIMATION OF CAR-FOLLOWING MODELS}
\label{sec:bayesian}
% What are car-following models and what are they good for?
Car-following models have been used extensively to model the behavior of human drivers when following a preceding vehicle in different traffic conditions. Car-following models have been shown to have variety of uses in different fields of transportation such as traffic flow (macroscopic) \cite{rothery1992car} and vehicle dynamics (microscopic) \cite{gunawan2012two}. These models aim at revealing interactions between an ego vehicle and its corresponding leading vehicle within the same lane, thus, creating the longitudinal profile of the ego vehicle over time. Individual values of the model parameters also reveal the driving characteristics. With increasing amount of naturalistic data of individual vehicles collected in different kinds of scenarios, there has been a larger focus towards calibrating car-following models from high accuracy vehicle trajectory data. 

% What previous work has been done in calibration of car-following models?
There has been a lot of research done on parameter estimation of car-following models from naturalistic datasets under different scenario types. Earlier research utilized macroscopic data to calibrate car-following models, which made the models less precise. Helbing and Tilch proposed the General Force model and calibrated it with Bosch GmbH car-following data collected using radars \cite{helbing1998generalized}. Kesting and Treiber calibrated and compared intelligent driver model (IDM) with Optimal Velocity model using the Bosch GmbH dataset using genetic algorithm \cite{kesting2008calibrating}. They were able to show that the errors between the actual trajectory data and the predicted values were below 30\% and the intra-driver variability accounts for the larger amount of error than the inter-driver variability. In the past decade, there has been an explosion of studies for calibration of car-following models using different kinds of naturalistic data and under different scenarios \cite{rahman2013application, treiber2013microscopic, papathanasopoulou2015towards}. 

% what do we plan to do?
In order to simulate real-world driving, we have to sample different parameter values from the distributions for each individual vehicle. We choose IDM as the car-following model in this research and estimate parameters using naturalistic driving data (NDD) from the Safety Pilot Model Deployment (SPMD) program \cite{bezzina2014safety}. Since we require the entire distribution to sample from, we employ Bayesian statistics in the form of Markov Chain Monte Carlo (MCMC) methods. MCMC methods employ sequential sampling using Markov Chain starting from a prior probability distribution and moving towards the desired distribution. We utilize a particular MCMC method, Metropolis-Hastings (M-H) algorithm, to determine parameter distributions for highway and urban scenarios using SPMD data. We then sample from these distributions to create background vehicles for the SUMO-Gym environment automatically. We describe the process in the subsections below. 

% what are the different kinds of car-following models?
\subsection{Intelligent Driver Model}
Car-following models have been developed from the 1950s onwards in order to understand the driving patterns of individual drivers \cite{brackstone1999car}. Numerous car-following models have been developed in the last seven decades such as the Gazis-Herman-Rothery (GHR) model \cite{gazis1961nonlinear}, IDM \cite{treiber2000congested}, the optimal velocity model \cite{bando1995dynamical}, and the models proposed by \cite{helly1959simulation}, \cite{gipps1981behavioural}, and \cite{wiedemann1974simulation}. Currently, the one of the most popular car-following models is the IDM based on the desired measures model which assumes that each individual driver has a desired following distance and following speed and the driver tries to minimize the gap between the actual measures and the desired. The acceleration for a vehicle $n$ is given as:

\begin{equation}
    a_n (t) = a_{max}^{(n)}\bigg(1 - \bigg(\frac{v_n(t)}{v_{des}(t)}\bigg)^\delta - \bigg(\frac{d_{des}(t)}{d_{front}(t)}\bigg)^2\bigg)
    \label{eq:acc_idm}
\end{equation}

where $a_{max}^{(n)}$ is the maximum acceleration of the vehicle, $v_n(t)$ is the actual speed of the vehicle, $v_{des}(t)$ is the desired speed of the vehicle, $d_{front}$ is the distance to the preceding vehicle and $d_{des}$ is the desired distance to the preceding vehicle. The desired following distance is dependent on various factors: speed of the vehicle ($v_n(t)$), speed difference ($\Delta v$), maximum acceleration ($a_{max}^{(n)}$), comfortable deceleration ($a_{comf}^{(n)}$), minimum time headway ($T$) and minimum distance to preceding vehicle ($d_{min}$). Mathematically, it can represented as:

\begin{equation}
    d_{des} = d_{min} + v_n(t) T + \frac{v_n(t) \Delta v}{2\sqrt{a_{max}^{(n)} a_{comf}^{(n)}}}
    \label{eq:dist_idm}
\end{equation}
The calibration of this model requires estimation of six parameters: $a_{max}^{(n)}$, $a_{comf}^{(n)}$, $v_{des}(t)$, $d_{min}$, $T$ and $\delta$, all of them should be positive. 

\subsection{Data collection and filtering}
SPMD was a project undertaken by the U.S. Department of Transportation (USDOT) to evaluate the effectiveness of vehicle-to-vehicle communication technology (V2V) in safety applications. University of Michigan Transportation Reseearch Institute (UMTRI) was the primary contractor and was in charge of device fitting on to vehicles of participant drivers, data collection and data management. The SPMD database is one of the largest databases in the world that recorded naturalistic driving behaviors over 34.9 million travel miles from 2842 equipped vehicles in Ann Arbor, Michigan. In the database, there are 98 sedans equipped with the data acquisition system (DAS). Since we only calibrate the car-following model in this research, we need the longitudinal data including the acceleration and speed of the ego-vehicle and the distance and speed difference to the preceding vehicle. As the first filtering step, we utilize the GPS data of fifty DAS-equipped vehicles to select portions of trajectories in highways and urban settings. This yields a large collection of trajectories in both bins which need to be further subsampled to feed into the M-H model. We filter out any start-and-stop trajectories and portions where lane changes occur from the trajectories. For the urban trajectories, we do not consider the trajectory portion near the intersections. 

\subsection{Bayesian Calibration}
The calibration for a car-following model based on trajectory data is essentially a non-linear optimization problem, in order to search for the best values of model parameters that minimize the measure of distance between the observed and the simulated ego-vehicle's behavior \cite{punzo2012can}. The optimization problem has the following form: 
\begin{equation}
    \theta^* = arg \min_{\theta \in D} f \bigg( Y^{obs}, Y(\theta)^{sim}\bigg)
    \label{eq:opt}
\end{equation}

where $\theta$ is the vector of parameters, $D$ is the domain of possible model parameters, $f$ is the distance function between observed ($Y^{obs}$) and computed acceleration ($Y(\theta)^{sim}$). 

From Equation~\ref{eq:acc_idm}, it is abundantly clear that it is a complex, nonstandard multivariate distribution. For such distributions, a popular method of calibration is utilizing Bayesian sequential sampling methods known as Markov Chain Monte Carlo (MCMC) methods. In this research, we utilize the M-H algorithm developed by Metropolis,  Rosenbluth, Rosenbluth, Teller, and  Teller  \cite{metropolis1953equation} and  subsequently  generalized  by  Hastings \cite{hastings1970monte}. The algorithm has two main ideas that combine together to provide a simple yet robust method - 
\begin{itemize}
    \item Transition function - at each step, there needs to be an update of parameter values i.e. given current parameter values, we need to calculate proposed parameter values. This is accomplished by provision of transition function. We utilize a very simple way of generating a new parameter values by adding a random noise term. The new parameter values are then calculated as $\theta_{prop} = \theta_{curr} + \xi$, where $\xi \backsim \mathcal{N}(0, \Sigma_{prop})$. The transition function is known as the random walk kernel and given as
    \begin{equation}
        Q(\theta_{prop} | \theta_{curr}) = \frac{1}{\sqrt{2\pi}} e^{-0.5(\theta_{prop} - \theta_{curr})^2}
        \label{eq:tran}
    \end{equation}
    \item Acceptance of proposed values - Given the proposed values, we have to decide whether to accept them as the new parameter values or sample again. This is done by calculating the acceptance probability:
    \begin{equation}
        \alpha = min \bigg( 1, \frac{\Pi(\theta_{prop})Q(\theta_{prop} | \theta_{curr})} {\Pi(\theta_{curr}) Q(\theta_{curr} | \theta_{prop})} \bigg)
    \end{equation}
    where $\Pi$ is the stationary distribution. The proposed parameter values are accepted with probability $\alpha$. Otherwise the current parameter values are retained. 
\end{itemize}

\subsection{Simulation results}
\begin{figure*}[thpb]
    \hspace{-1cm}
    \centering
    \includegraphics[scale=1.0]{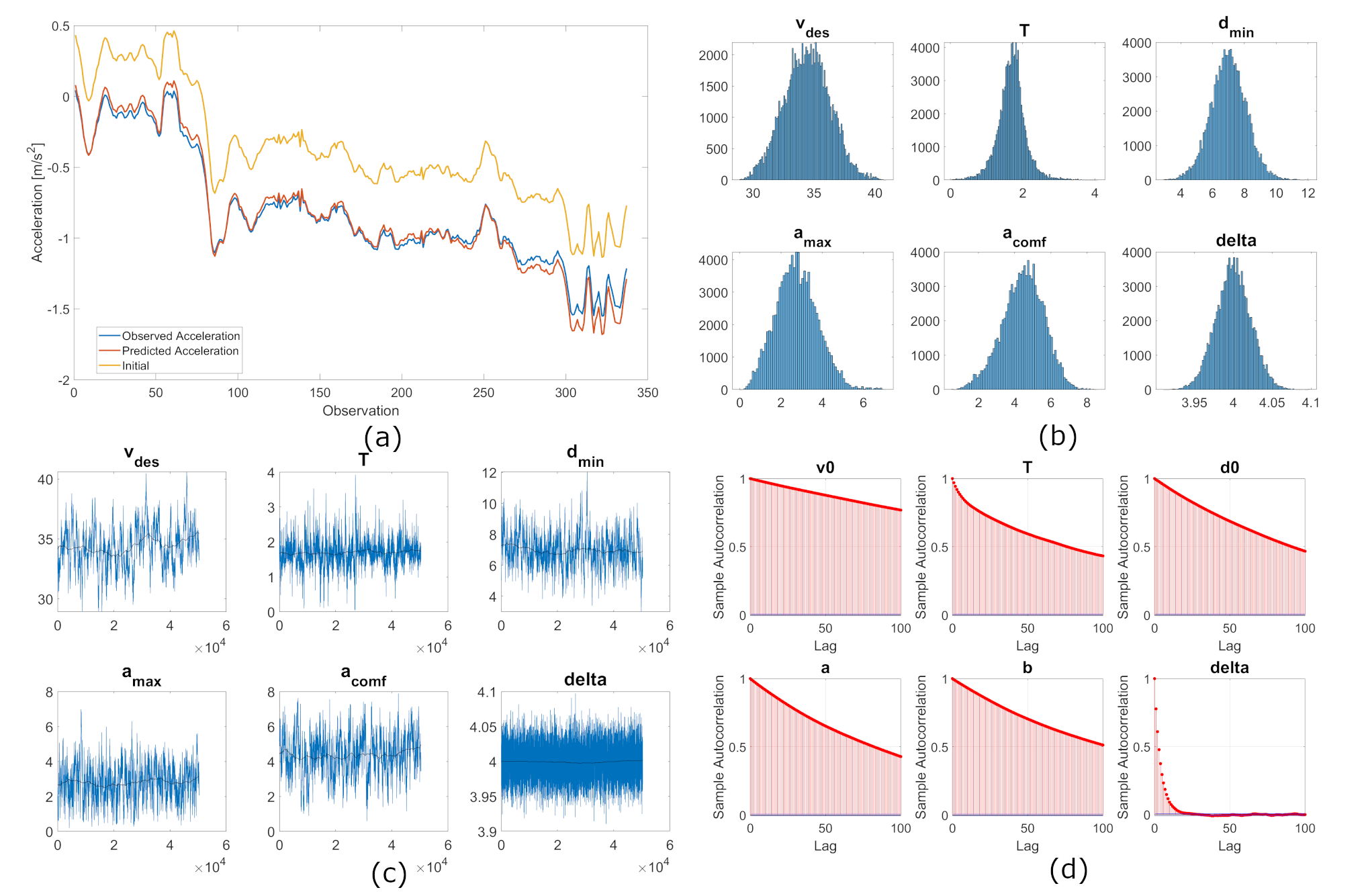}
    \caption{IDM parameter calibration using simulation data - (a) observed vs estimated acceleration; (b) parameter distributions; (c)parameter sampling over iterations; (d) autocorrelation between parameters}
    \label{fig:idm_sim}
\end{figure*}

We collect simulation data for 50 vehicles with 200 observations each with all vehicles running IDM with the following parameters: $a_{max} = 3 m/s^2$, $a_{comf} = 5 m/s^2$, $v_{des} = 35 m/s$, $d_{min} = 10 m$, $T = 2 s$ and $\delta = 4$. Fig.~\ref{fig:idm_sim} shows the results for the simulation using M-H algorithm. Fig.~\ref{fig:idm_sim}(a) provides the initial estimate of acceleration for an individual vehicle and the final estimate after running M-H algorithm for 100,000 iterations. We find that the M-H algorithm can approximate the observed acceleration closely, with the maximum deviation of 5\%. The final distributions of each parameter is given in Fig.~\ref{fig:idm_sim}(b). We find that the M-H algorithm can approximate the actual parameters very closely with the means of the distributions lying closely to the actual parameters. In order to understand the convergence of the Markov chains, we plot the samples over the iterations as given in Fig.~\ref{fig:idm_sim}(c). The M-H Markov chains are considered convergent if they samples look like random noise about a straight line. Finally, the autocorrelation plot in Fig.~\ref{fig:idm_sim}(d) shows values of the lag-k autocorrelation function (ACF) against increasing k values. The autocorrelation is an important diagnostic for convergence of MCMC chains and the lesser the correlation the better the convergence. Thus, we can show that M-H algorithm works really well to approximate the original values and can provide meaningful distributions. 

\subsection{Real data results}
\begin{figure}[thpb]
    \hspace{-1cm}
    \centering
    \includegraphics[scale=.13]{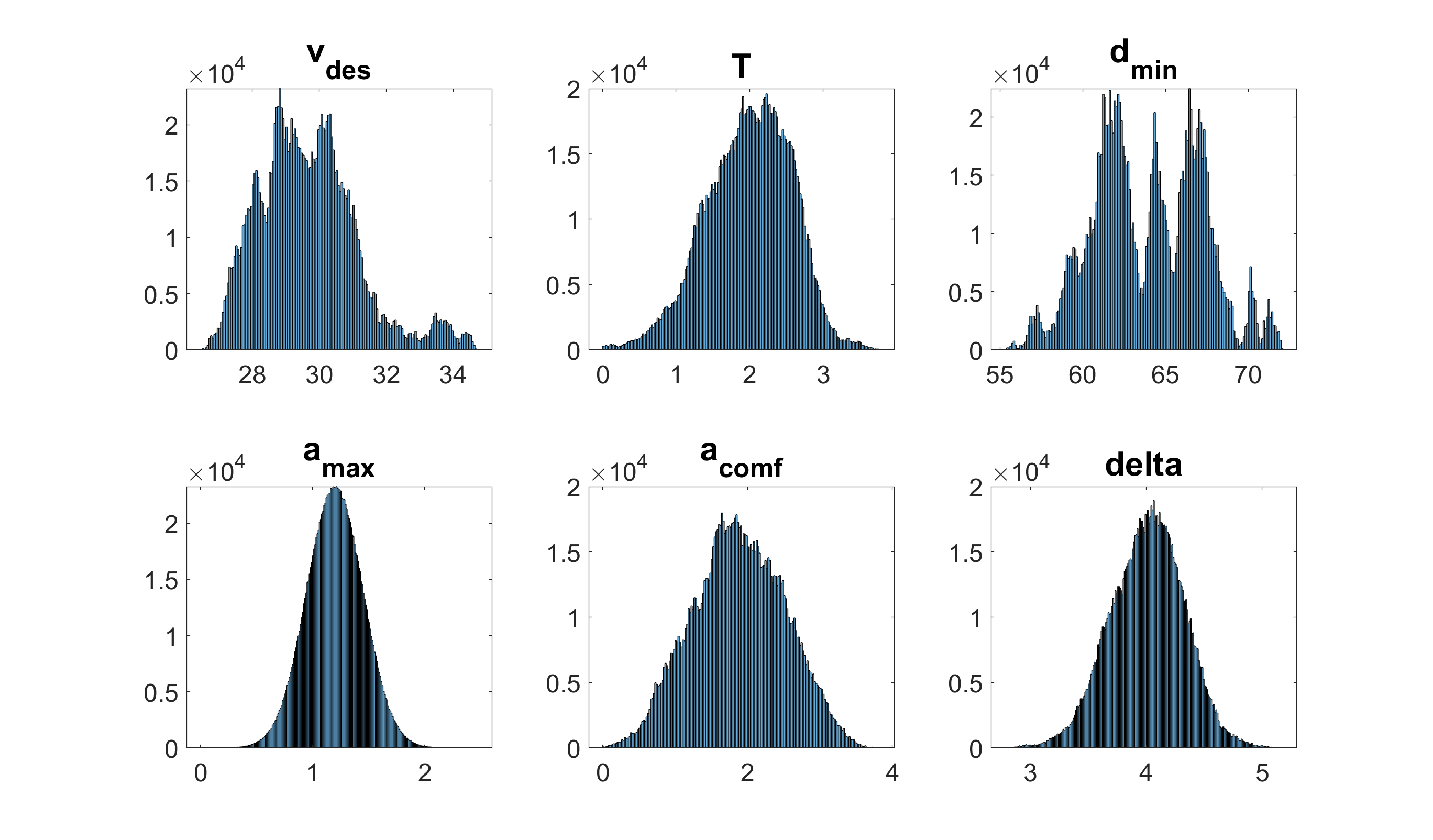}
    \caption{Parameter distributions from SPMD highway data}
    \label{fig:param_highway}
\end{figure}

\begin{figure}[thpb]
    \hspace{-1cm}
    \centering
    \includegraphics[scale=.13]{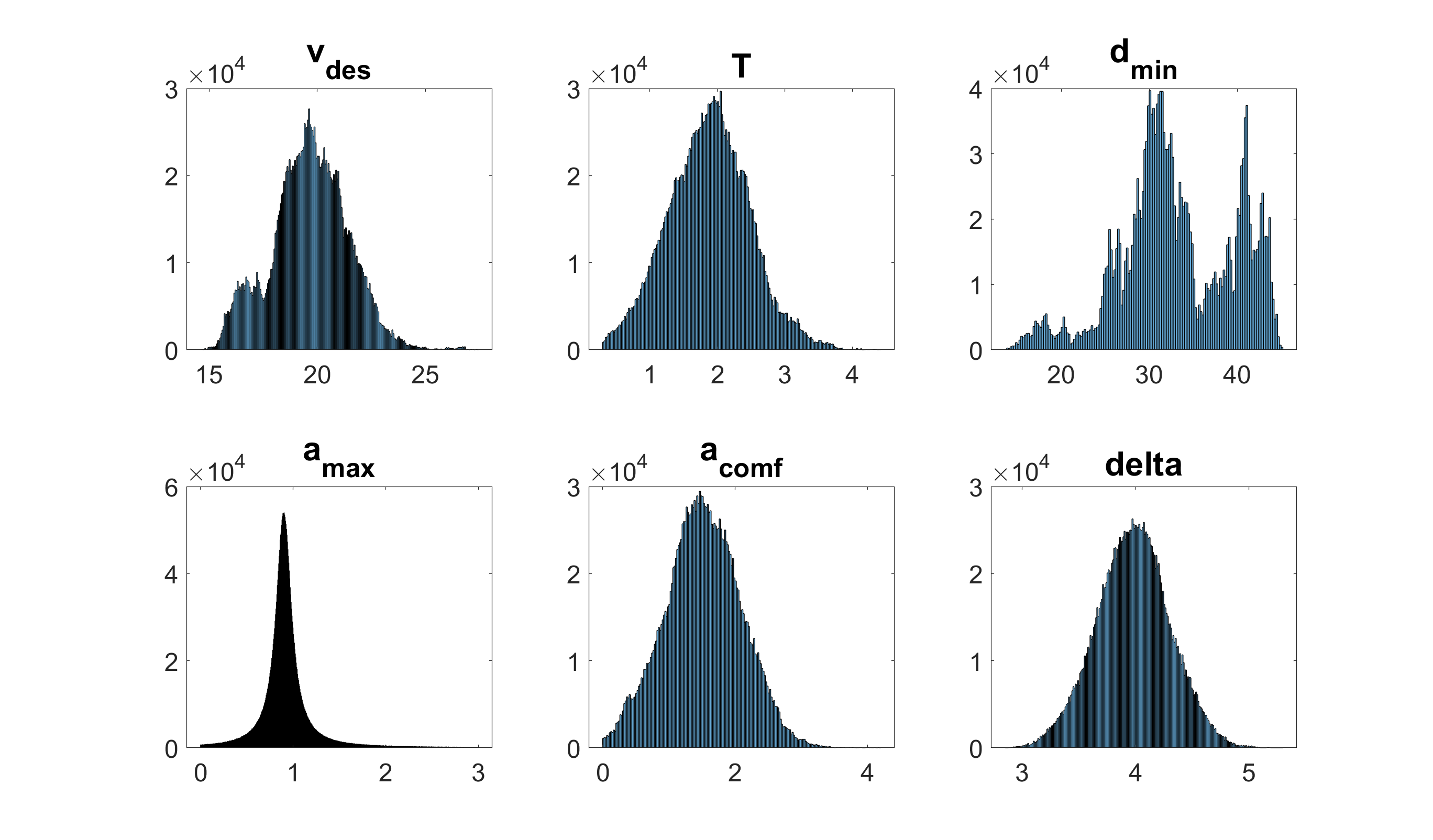}
    \caption{Parameter distributions from SPMD urban data}
    \label{fig:param_urban}
\end{figure}

We present the distribution results from running the M-H algorithm on the SPMD highway and urban data in Fig.~\ref{fig:param_highway} and Fig.~\ref{fig:param_urban}. It is evident from the plots, that the desired speed in highway is an unimodal distribution centered around 30 m/s, whereas in the urban data, it is a bimodal distribution with two peaks centered on 18 and 20 m/s. Another large difference is in the minimum distance, where the ranges lie between 65-75 m in highway driving data while they lie between 15-45 m in the urban data. We provide a comparison table (Table~\ref{table:comparison}) between the mean values found by us to the ones found in the literature. From the table, we see that while there is agreement between the values of desired speed,  acceleration and $\delta$, the values vary for other parameter values. 

\begin{table}[ht] 
\caption{Comparison of values between our method and given in \cite{schakel2012integrated} for highways} % title of Table 
\centering      % used for centering table 
\begin{tabular}{c c c}  % centered columns (3 columns) 
\hline                       %inserts double horizontal lines 
Parameter values & Our method & Schakel et. al.\cite{schakel2012integrated} \\ [0.5ex] % inserts table 
%heading 
\hline                    % inserts single horizontal line 
$v_{des} [m/s]$  & 29.7 & 34.4 \\    % inserting body of the table 
$T [s]$ & 2.0 & 1.2  \\ 
$d_{min} [m]$ & 63.9 & 7.0  \\ 
$a_{max} [m/s^2]$ & 1.2 & 1.0 \\ 
$a_{comf} [m/s^2]$ & 2.0 & 1.67 \\
$\delta$ & 4.0 & 4.0 \\ [1ex]       % [1ex] adds vertical space 
\hline     %inserts single line 
\end{tabular} 
\label{table:comparison}  % is used to refer this table in the text 
\end{table}

% \begin{figure*}[thpb]
%     \centering
%     \includegraphics[scale=.25]{Figures/parameter_IDM_time_series_highway.png}
%     \caption{Time series of parameter estimation from SPMD highway data with moving average shown as a black line}
%     \label{fig:time_series_highway}
% \end{figure*}

% \begin{figure*}[thpb]
%     \centering
%     \includegraphics[scale=.25]{Figures/parameter_IDM_time_series_urban.png}
%     \caption{Time series of parameter estimation from SPMD urban data with moving average shown as a black line}
%     \label{fig:time_series_urban}
% \end{figure*}

\subsection{Assimilating into SUMO-Gym}
Given the parameter distributions for highway and urban scenarios, we need to sample parameters for individual vehicles from the output distributions. Since most of these parameters have multimodal distributions, we sample directly from the histograms. We sample by randomly selecting a bin according to its probability mass and then sampling a value from an uniform distribution over the chosen bin. These are the parameter values that we input into the SUMO-Gym environment. We create routes using randomTrips.py script in SUMO which generates a set of random trips for a given network. We use an automated script to place a certain number of vehicles in each route with parameter values sampled from the parameter distribution file as described above. The script outputs a route file which is used by the SUMO-Gym environment. 

\section{SCENARIO ABSTRACTION COMBINING SUMO AND OPENAI GYM}
\label{sec:scenario}
% First have to talk about the problem of installing and setting up a simple scenario in SUMO - that will provide the motivation
% Introduction of OpenAI Gym
\subsection{Introduction to OpenAI Gym}
\begin{figure}[thpb]
    \centering
    \includegraphics[scale=.42]{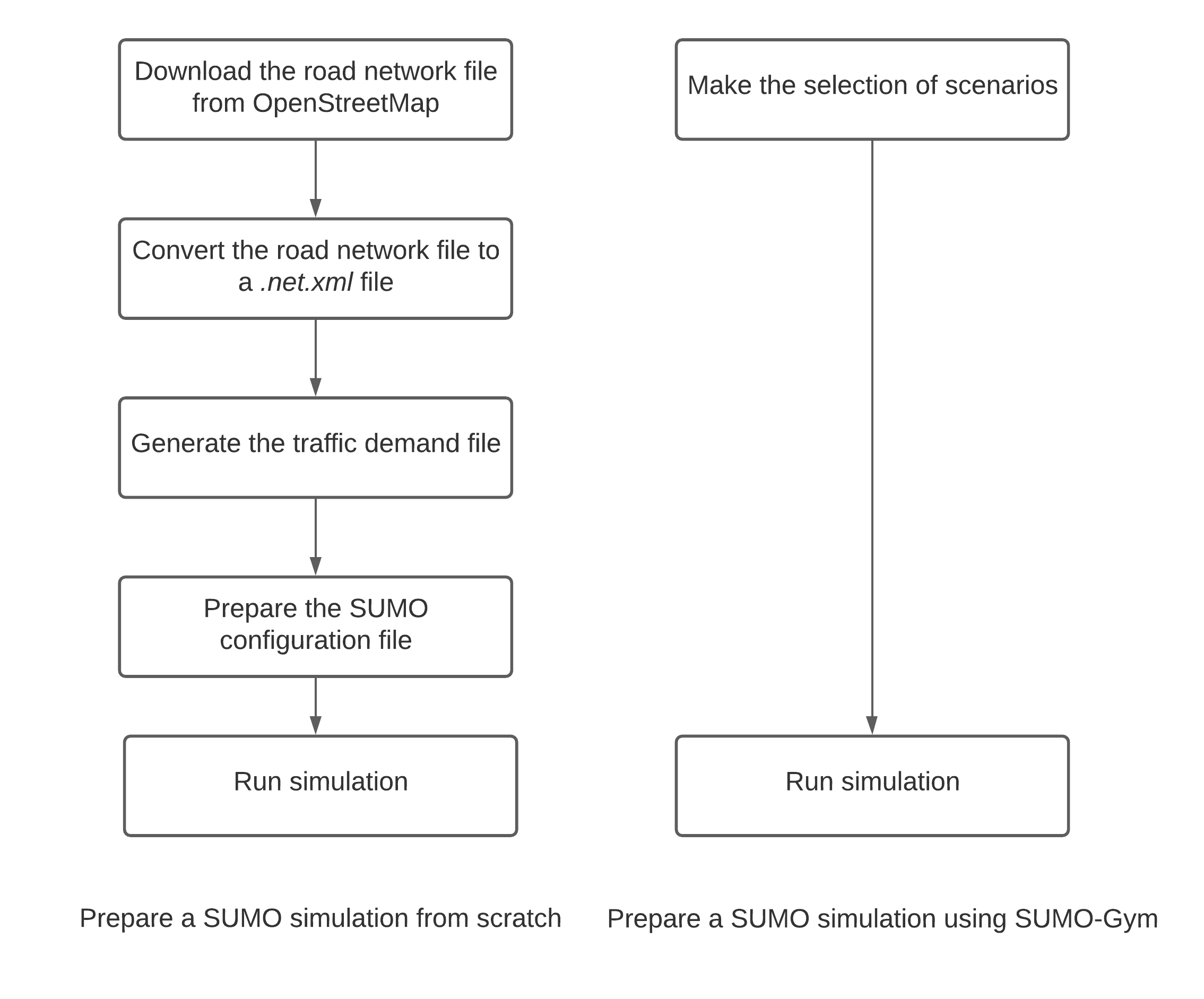}
    \caption{Comparison of developing a traffic simulation from scratch and using SUMO-Gym  }
    \label{fig:compare_charts}
\end{figure}

Gym is an open source Python library developed by OpenAI \cite{1606.01540}. It was originally designed for developing and testing reinforcement learning algorithms. Gym provides a convenient way to communicate between algorithms and environments by using its standard API by creating a common interface for all the simulation environments and abstracting the process of controlling the agents in the simulation via easy-to-use commands. 
% why combine with OpenAI gym - what are the benefits - simple easy to use commands, actual environment abstracted and the end user doesnt need to interact with environment directly - helps in our case as well
This paper aims to leverage the power of OpenAI Gym to abstract the SUMO simulation environment and propose a novel AV testing platform named SUMO-Gym. While SUMO users need to create a road network file and a route file containing the dynamic traffic elements for each new simulation, Similar to other environments of OpenAI Gym, users of SUMO-Gym only need to prepare their AV pipeline without setting up the simulation environment. Fig.~\ref{fig:compare_charts} compares the process of preparing a SUMO simulation from scratch to that using SUMO-Gym. 

%Concept of SUMO-Gym
\subsection{Introduction to SUMO-Gym}
\begin{figure}[thpb]
    \centering
    \includegraphics[scale=.38]{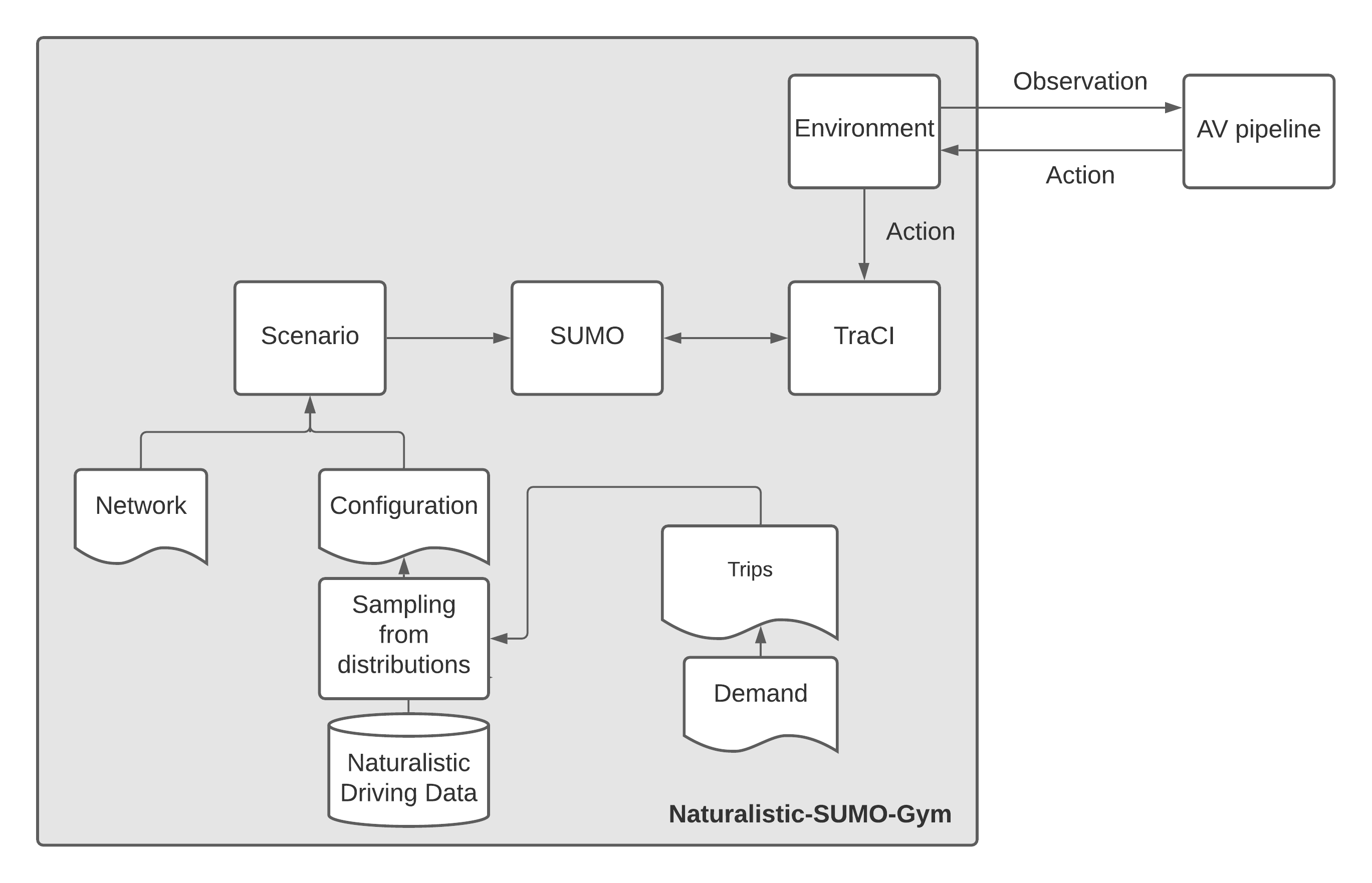}
    \caption{Components of SUMO-Gym}
    \label{fig:flowchart}
\end{figure}

SUMO-Gym is a portable, user-friendly AV simulation platform. Users of SUMO-Gym can test their AV pipeline to generate actions compatible with the package based on generic observations from SUMO-Gym. Fig.~\ref{fig:flowchart} shows the interactions between different components of SUMO-Gym. The first part of SUMO-Gym is the simulation scenario. SUMO-Gym provides multiple simulation scenarios which cover different ODDs, including highways, major arterials, and local collectors. % have to talk about the real world layouts being provided in the package with the possibility of adding others

\subsection{Scenario abstraction with SUMO-Gym}
\begin{figure}[thpb]
    \centering
    \includegraphics[scale=.17]{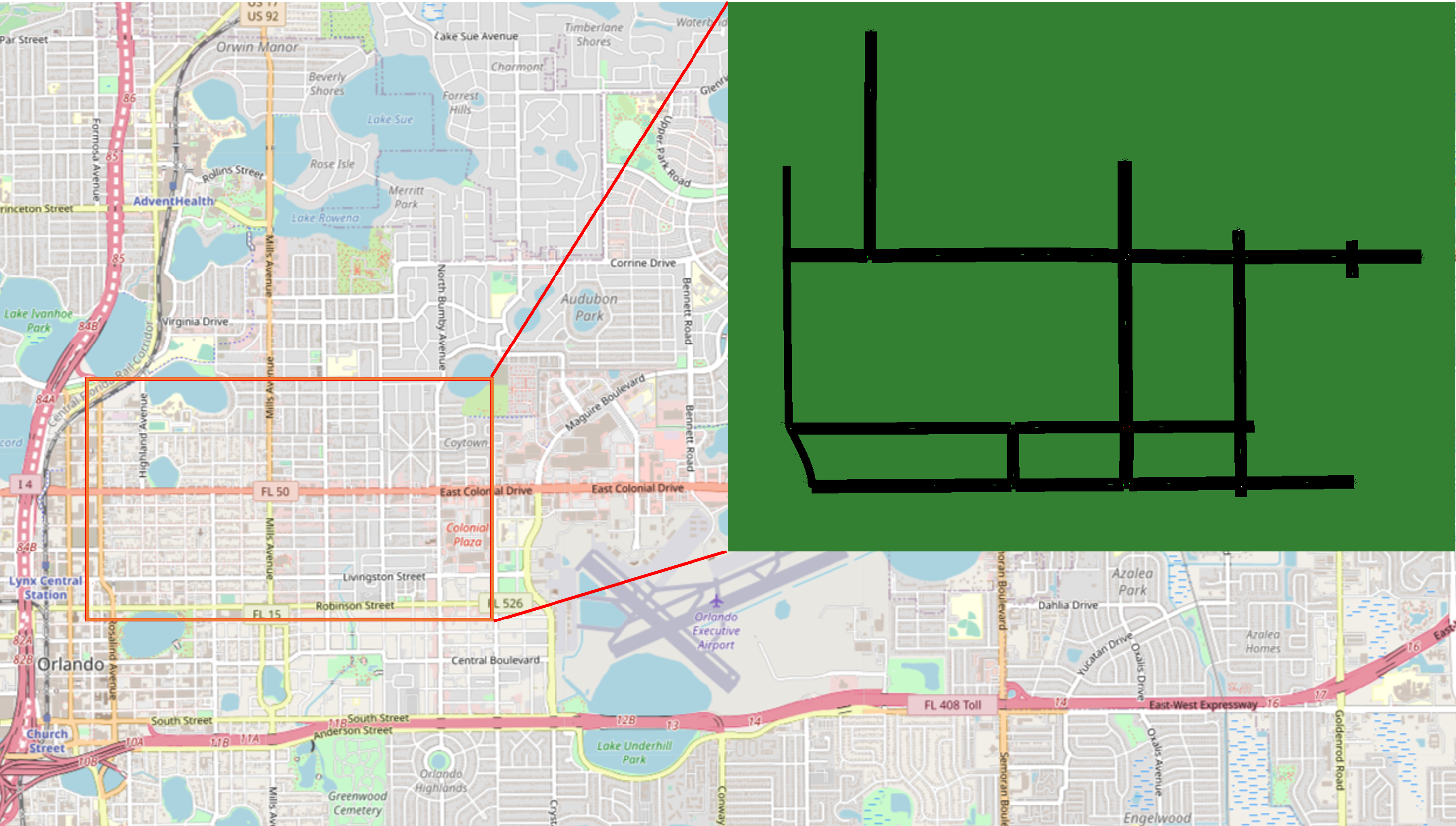}
    \caption{Example of an urban scenario layout}
    \label{fig:urban_scenario}
\end{figure}

SUMO-Gym lets users choose preset scenario types (such as highway and urban). For each type of the scenarios, we create a library of road network layouts and dynamic route files, and the subsequent configuration files. A random configuration file of the scenario type is chosen and loaded by SUMO-Gym.  Two types of scenarios were included in SUMO-Gym, highway and urban scenario. For example, the road network of highways around Ann Arbor was used to develop an highway scenario while the traffic network of downtown Orlando was used to develop an urban scenario. The layout along with the OSM map is shown in Fig.~\ref{fig:urban_scenario}. In addition, SUMO-Gym has the ability to operate using customized traffic scenarios based on users' needs. In general, a simulation scenario has three crucial files: network, demand, and configuration file. Specifically, the network file defines the road network of the simulation. The demand file consists of various traffic routes, which contain vehicles' trajectory information. Moreover, to make the demand file can reflect realistic driving environment, parameters of the car-following model was calibrated using naturalistic driving data. A configuration file is then used to provide network and demand files as inputs to the SUMO simulation. The testing process is then run based on the particular configuration file. 

\subsection{Inheriting environment from Gym}
The second component of SUMO-Gym is the environment, a Python class which inherits from Gym and provides a convenient way to run SUMO simulation with few generic commands. The environment contains multiple Python functions to initialize, reset, render, start, step through, and stop a simulation. The environment uses TraCI~\cite{wegener2008traci} to control and communicate with SUMO. The environment provides observations currently in the form of an observation matrix to users of SUMO-Gym. The observation matrix is a $V \times F$ array for $V$ neighboring vehicles (leaders and followers to the ego vehicle in each lane) and $F$ features: presence, x-position, y-position, velocity in x and velocity in y. Users of SUMO-Gym can use their customized AV algorithms to generate actions for the environment. Currently, we utilize continuous action in the form of longitudinal and lateral acceleration at each step. Note that the clipping of acceleration is to be performed by the user before inputting into SUMO-Gym. We compute the change in x and y positions of the ego vehicle in the road aligned coordinate system and transform them to SUMO's global coordinate system. The ego vehicle is then moved via the moveToXY function in TraCI. During each step, we also check whether the ego vehicle has veered off the road or is in a collision. In such cases, we stop the simulation and send pertinent information back to the user. 

\begin{figure}[thpb]
    \centering
    \includegraphics[scale=.46]{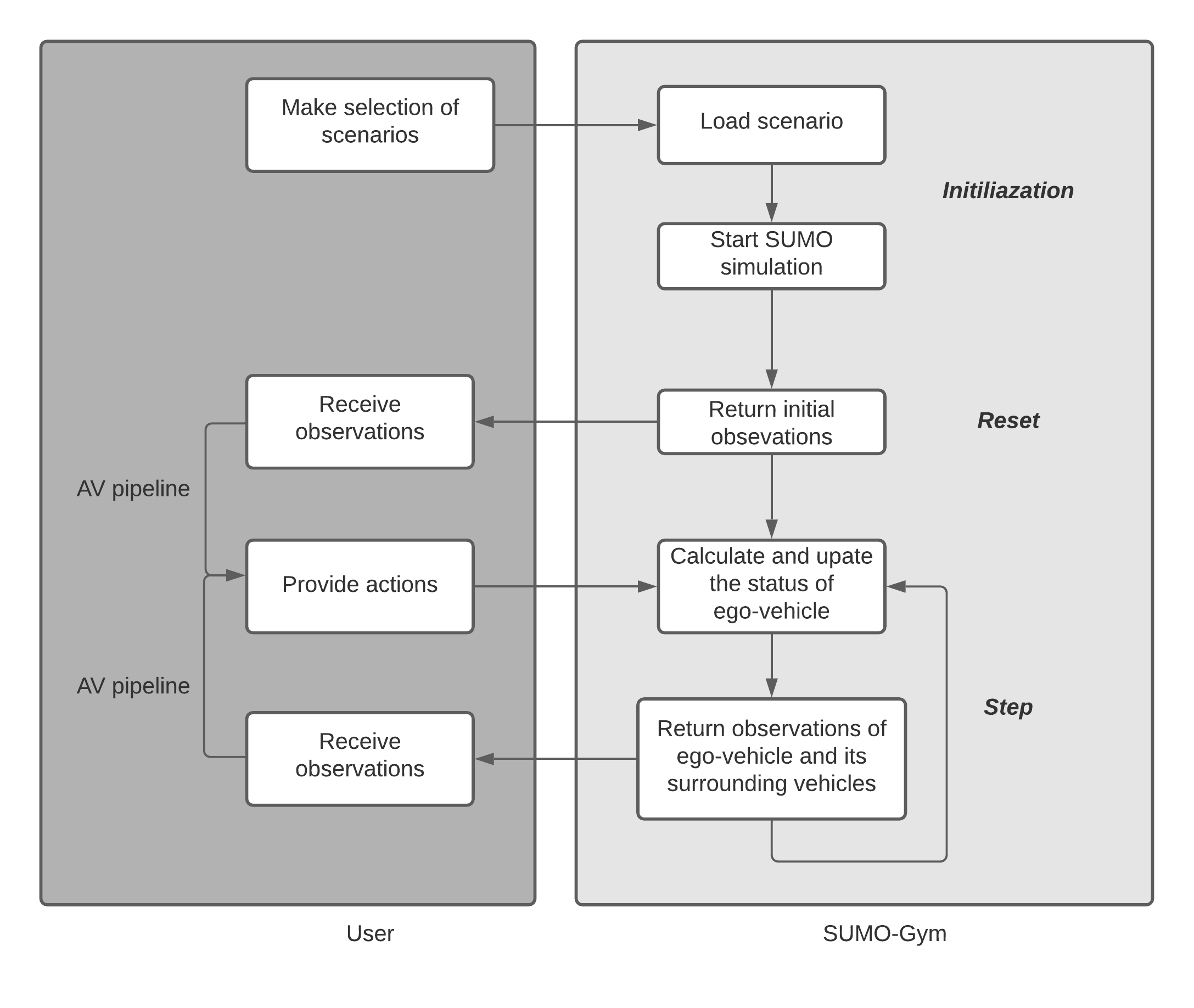}
    \caption{SUMO-Gym diagram}
    \label{fig:diagram}
\end{figure}

Fig.~\ref{fig:diagram} provides detailed information about the process diagram of SUMO-Gym. Users of SUMO-Gym need to make the selection of scenarios and pass this information to SUMO-Gym. SUMO-Gym first loads the scenario and start the simulation as the initialization. Then, SUMO-Gym resets the simulation and returns initial observations to users. Using actions provided by users' AV pipeline, SUMO-Gym will iteratively take a simulation step, which updates the status of ego-vehicle and returns observations. In the current framework, we have a placeholder for reward function. If so desired, the user can modify the function and receive a scalar reward at each simulation step. %It may be helpful to add equations of longitudinal and lateral acceleration, please advise. 
In addition, SUMO-Gym can render the simulation using SUMO-GUI, which helps users visually observe the simulation.

\section{DISCUSSION AND CONCLUSIONS}
\label{sec:discussion}
% What we need to talk about: 

% 1. how these improve the final results?
% 2. real-world value of these enhancements for the practitioners?
% 3. future work
We present a software solution for AV based simulation and testing geared towards user convenience and providing real world variability. We provide the software as an open source package available as a docker container \url{https://github.com/arpan-kusari/Naturalistic-SUMO-Gym}. The docker container installs SUMO, OpenAI Gym and SUMO-Gym package. In addition to the SUMO-Gym package, we also include a Python script with the ego vehicle running a simple IDM model to test the package. By abstracting the entire simulation environment, SUMO-Gym makes SUMO more user-friendly and lets users solely focus on preparing their AV pipeline. Our hope in releasing this software is to provide a platform which can accelerate the process of verification and validation of autonomous vehicles. 

In terms of future work, we would like to increase the number of scenarios in each scenario type and increase functionality of the simulation environment. Specifically, we would like to broaden the type of observations and actions that are inputted and outputted by the simulation. Our other aim in releasing this software is to solicit opinions from the community at large in terms of features and functionalities that could be incorporated into the package. 

\bibliographystyle{IEEEtran}
\bibliography{IEEEfull}

\end{document}